\documentclass[11pt]{article}
\usepackage[utf8]{inputenc}
\usepackage[margin=1in]{geometry}
\usepackage[T1]{fontenc} %
\usepackage{graphicx}
\usepackage{mathpazo}
\usepackage{hyperref}
\hypersetup{
  colorlinks = true,  
  urlcolor = {blueGrotto},
  linkcolor = {royalBlue},
  citecolor = {navyBlue}
}
\usepackage{booktabs} 
\usepackage{multirow} 
\usepackage{multicol} 
\usepackage[]{color-edits} %

\addauthor[Grigoris]{gv}{teal}

\addauthor[Omar]{om}{blue}

\usepackage{xcolor}
\definecolor{niceRed}{RGB}{190,38,38}
\definecolor{blueGrotto}{HTML}{059DC0}
\definecolor{royalBlue}{HTML}{057DCD}
\definecolor{navyBlue}{HTML}{0B579C}
\definecolor{limeGreen}{HTML}{81B622}
\definecolor{nicePurple}{HTML}{9c27b0}
\definecolor{lightRoyalBlue}{HTML}{def2ff} 
 \definecolor{gold}{HTML}{ffa300}

\usepackage[
  backref=true,
  backend=biber,
  natbib=true,
  style=alphabetic,
  sorting=alphabeticlabel,
  sortcites=true,
  minbibnames=3,
  maxbibnames=999,
  mincitenames=1,
  maxcitenames=5,
  minalphanames=3,
  maxalphanames=5
]{biblatex}

\DeclareSortingTemplate{alphabeticlabel}{
  \sort[final]{%
    \field{labelalpha} %
  }
  \sort{%
    \field{year}   %
  }
  \sort{%
    \field{title}  %
  }
}

\AtBeginRefsection{\GenRefcontextData{sorting=ynt}}
\AtEveryCite{\localrefcontext[sorting=ynt]}
\usepackage[utf8]{inputenc}

\usepackage{amsmath}
\usepackage{amssymb}
\usepackage{amsthm}
\usepackage{bbm}
\usepackage{blkarray}
\usepackage{color}
\usepackage{enumerate}
\usepackage{float} 
\usepackage{subcaption}
\usepackage{tikz}
\usetikzlibrary{arrows.meta, positioning, calc, shapes, fit}
\usepackage{hyperref}
\usepackage[capitalise,noabbrev,nameinlink]{cleveref} %
\usepackage{mathtools}
\usepackage[inline]{enumitem} 
\usepackage{tcolorbox}
\usepackage{nicefrac}
\usepackage{dsfont}
\usepackage[framemethod=TikZ]{mdframed}
\usepackage{todonotes}
\usepackage{changepage} %
\usepackage{pifont} %
\usepackage{thm-restate} %
\usepackage{centernot}

\tikzset{
  myNodeFlex/.style={
    draw,
    rectangle,
    rounded corners,
    text centered,
    minimum height=1.5em,
  }
}

\tikzset{
  myNode/.style={
    draw,
    rectangle,
    rounded corners,
    text centered,
    minimum height=1.5em,
    minimum width=3cm,
    text width=5cm,    
  }
}

\tikzset{
  myNodeNarrow/.style={
    draw,
    rectangle,
    rounded corners,
    text centered,
    minimum height=1.5em,
    minimum width=1cm,
  }
}

\tikzset{
  myNodeWide/.style={
    draw,
    rectangle,
    rounded corners,
    text centered,
    minimum height=1.5em,
    minimum width=6cm,
  }
}

\usepackage{algorithm}
\usepackage{algpseudocode}[1]
\renewcommand{\algorithmicrequire}{\textbf{Input:}}

\usepackage{mathrsfs} %
\usepackage{soul} %

\theoremstyle{plain} %
\newtheorem{theorem}{Theorem}[section]
\newtheorem{corollary}[theorem]{Corollary}

\newtheorem{lemma}[theorem]{Lemma}

\newtheorem{definition}{Definition}

\newtheorem*{definition*}{Definition}

\theoremstyle{definition} %
\newtheorem{example}{Example}

\theoremstyle{remark} %

\AfterEndEnvironment{definition}{\noindent\ignorespaces}
\AfterEndEnvironment{infdefinition}{\noindent\ignorespaces}
\AfterEndEnvironment{example}{\noindent\ignorespaces}
\AfterEndEnvironment{assumption}{\noindent\ignorespaces}
\AfterEndEnvironment{lemma}{\noindent\ignorespaces}
\AfterEndEnvironment{theorem}{\noindent\ignorespaces}
\AfterEndEnvironment{proposition}{\noindent\ignorespaces}
\AfterEndEnvironment{fact}{\noindent\ignorespaces}
\AfterEndEnvironment{question}{\noindent\ignorespaces}
\AfterEndEnvironment{corollary}{\noindent\ignorespaces}
\AfterEndEnvironment{model}{\noindent\ignorespaces}
\AfterEndEnvironment{remark}{\noindent\ignorespaces}
\AfterEndEnvironment{proof}{\noindent\ignorespaces}
\AfterEndEnvironment{fact}{\noindent\ignorespaces}
\AfterEndEnvironment{minftheorem}{\noindent\ignorespaces}
\AfterEndEnvironment{inftheorem}{\noindent\ignorespaces}
\AfterEndEnvironment{maintheorem}{\noindent\ignorespaces}
\AfterEndEnvironment{restatable}{\noindent\ignorespaces}
\AfterEndEnvironment{observation}{\noindent\ignorespaces}

\crefname{section}{Section}{Sections}
\crefname{theorem}{Theorem}{Theorems}
\crefname{theorem*}{Theorem}{Theorems}
\crefname{inftheorem}{Informal Theorem}{Informal Theorems}
\crefname{assumption}{Assumption}{Assumptions}
\crefname{lemma}{Lemma}{Lemmas}
\crefname{definition}{Definition}{Definitions}
\crefname{infdefinition}{Informal Definition}{Informal Definitions}
\crefname{conjecture}{Conjecture}{Conjectures}
\crefname{corollary}{Corollary}{Corollaries}
\crefname{construction}{Construction}{Constructions}
\crefname{conjecture}{Conjecture}{Conjectures}
\crefname{claim}{Claim}{Claims}
\crefname{observation}{Observation}{Observations}
\crefname{proposition}{Proposition}{Propositions}
\crefname{fact}{Fact}{Facts}
\crefname{question}{Question}{Questions}
\crefname{problem}{Problem}{Problems}
\crefname{remark}{Remark}{Remarks}
\crefname{example}{Example}{Examples}
\crefname{equation}{Equation}{Equations}
\crefname{appendix}{Appendix}{Appendices}
\crefname{algorithm}{Algorithm}{Algorithms}
\crefname{model}{Model}{Models}
\crefname{figure}{Figure}{Figures}
\crefname{condition}{Condition}{Conditions}

\usepackage{etoolbox}

\BeforeBeginEnvironment{subenvironment}{\white{.}\\ \vspace{-10mm}}
\AfterEndEnvironment{subenvironment}{\vspace{4mm}}

\newcommand{\eat}[1]{}

\makeatletter
\newcommand{\tagnum}[2]{%
    \refstepcounter{equation}%
    \tag{#1) \ (\theequation}%
    \protected@write \@auxout {}{%
        \string \newlabel {#2}{{\theequation}{\thepage}{}{equation.\theequation}{}}%
    }%
}
\makeatother

\newcommand{\white}[1]{\textcolor{white}{#1}}

\newcommand{\inbrace}[1]{\left\{#1\right\}}

\newcommand{\inparen}[1]{\left(#1\right)}

\newcommand{\N}{\mathbb{N}}

\newcommand\ind{\mathds{1}}

\renewcommand{\epsilon}{\varepsilon}

\makeatletter
\newcommand*{\tran}{{\mathpalette\@tran{}}}
\newcommand*{\@tran}[2]{\raisebox{\depth}{$\m@th#1\intercal$}}
\makeatother

\def\<{\langle}
\def\>{\rangle}

\DeclareMathAlphabet{\mathpzc}{OT1}{pzc}{m}{it}

\newcommand{\customcal}[1]{\euscr{#1}}

\newcommand{\cC}{\customcal{C}}

\newcommand{\cL}{\customcal{L}}

\newcommand{\cN}{\customcal{N}}

\newcommand{\cX}{\customcal{X}}

\DeclareMathAlphabet{\mathdutchcal}{U}{dutchcal}{m}{n}
\SetMathAlphabet{\mathdutchcal}{bold}{U}{dutchcal}{b}{n}
\DeclareMathAlphabet{\mathdutchbcal}{U}{dutchcal}{b}{n}
 
\DeclareMathAlphabet\urwscr{U}{urwchancal}{b}{n}%
\DeclareMathAlphabet\rsfscr{U}{rsfso}{m}{n}
\DeclareMathAlphabet\euscr{U}{eus}{m}{n}
\DeclareFontEncoding{LS2}{}{}
\DeclareFontSubstitution{LS2}{stix}{m}{n}
\DeclareMathAlphabet\stixcal{LS2}{stixcal}{m} {n}

\renewcommand{\paragraph}[1]{\bigskip \noindent\textbf{#1}~~}

\newcommand{\ie}{\textit{i.e.}}

        \usepackage{tikz}
\usetikzlibrary{patterns}

\newcommand{\algo}[1]{\mathpzc{#1}}
\newcommand{\generator}{\algo{G}}

\renewcommand{\hat}{\widehat}

\usepackage{array}
\newcolumntype{L}[1]{>{\raggedright\let\newline\\\arraybackslash\hspace{0pt}}m{#1}}
\newcolumntype{C}[1]{>{\centering\let\newline\\\arraybackslash\hspace{0pt}}m{#1}}
\newcolumntype{R}[1]{>{\raggedleft\let\newline\\\arraybackslash\hspace{0pt}}m{#1}}

\newmdenv[
    backgroundcolor=lightgray!10, %
    roundcorner=5pt,              %
    linecolor=black,              %
    linewidth=1pt,                %
    innertopmargin=6pt,           %
    innerbottommargin=6pt,        %
    innerleftmargin=5pt,          %
    innerrightmargin=5pt,         %
    skipabove=2pt,                %
    skipbelow=0pt                 %
]{curvybox}

\title{(Im)possibility of Automated Hallucination Detection in Large Language Models}

\author{
\begin{tabular}{cc}
    \begin{tabular}{c}
        \textbf{Amin Karbasi} \\
        Yale University \\
        \texttt{amin.karbasi@yale.edu}
    \end{tabular}
    &
    \begin{tabular}{c}
        \textbf{Omar Montasser} \\
        Yale University \\
        \texttt{omar.montasser@yale.edu}
    \end{tabular}
    \\
    \\
    \begin{tabular}{c}
        \textbf{John Sous} \\
        Yale University \\
        \texttt{john.sous@yale.edu}
    \end{tabular}
    &
    \begin{tabular}{c}
        \textbf{Grigoris Velegkas} \\
        Yale University \\
        \texttt{grigoris.velegkas@yale.edu}
    \end{tabular}
\end{tabular}
}

\date{}

\begin{document}

\maketitle
    
\begin{abstract}

Is automated hallucination detection fundamentally possible?
In this work, we introduce a theoretical framework to rigorously analyze the feasibility of automatically detecting hallucinations produced by large language models (LLMs). Inspired by the classical Gold-Angluin framework for language identification \citep{gold1967language,angluin1979finding,angluin1980inductive} and its recent adaptation to language generation by \citet{kleinberg2024language}, we investigate whether an algorithm—trained on examples drawn from an unknown target language $K$ selected from a countable collection $\cL$, and provided access to an LLM—can reliably determine if the LLM’s outputs are correct or constitute hallucinations.

First, we establish an equivalence between hallucination detection and the classical task of language identification. We prove that any successful hallucination detection method can be converted to a language identification method, and conversely, algorithms solving language identification can be  used for hallucination detection. Given the known inherent difficulty of language identification, our result implies that automated hallucination detection is fundamentally \emph{impossible} for most language collections if the detector is trained using only correct examples (positive examples) from the target language.

Second, we demonstrate that the use of expert-labeled feedback—i.e., training the detector with both positive examples (correct statements) and negative examples (explicitly labeled incorrect statements)—dramatically alters this conclusion. Under this enriched training regime, automated hallucination detection becomes \emph{possible} for \emph{all} countable language collections.

These results highlight the essential role of expert-labeled examples in the training process of hallucination detectors. Thus, they provide theoretical support for feedback-based methodologies, such as reinforcement learning with human feedback (RLHF), further explaining why these approaches have proven indispensable in enhancing the reliability and safety of real-world LLM deployments.

\end{abstract}
\newpage

\section{Introduction}

The recent breakthroughs in Large Language Models (LLMs) have significantly advanced the state-of-the-art in natural language processing and broader machine learning tasks \citep{achiam2023gpt,team2024gemini}. Contemporary models routinely demonstrate exceptional performance across diverse tasks, including mathematical reasoning, complex problem-solving, and generating coherent, contextually appropriate text \citep{bubeck2023sparks,touvron2023llama}.

However, alongside these remarkable capabilities, a critical limitation has emerged: LLMs frequently produce \emph{hallucinations}—outputs that appear fluent and convincing yet are factually incorrect \citep{ji2023survey}. Hallucinations significantly limit the trustworthiness of LLMs, posing substantial risks when deploying them in sensitive applications, and raising urgent concerns around ethics, reliability, and societal impacts \citep{weidinger2021ethical,zhang2023siren,azamfirei2023large}.

A promising approach to addressing hallucinations is the development of automated detection mechanisms. Unfortunately, practical attempts to detect hallucinations using LLMs themselves as detectors have faced limitations. Empirical studies indicate that LLMs perform significantly worse than humans at identifying hallucinations, and typically require reliable external feedback—such as explicit labeling by experts—to improve \citep{kamoi2024evaluating,kamoi2024can}. Despite these empirical observations, a theoretical understanding of these practical difficulties has remained open:

\begin{center} \emph{Is automated hallucination detection inherently difficult, or can we expect it to become easier as models improve?} \end{center}

To address this gap, we introduce a formal theoretical framework inspired by classical learning theory—particularly the foundational work of Gold and Angluin on \emph{language identification} \citep{gold1967language,angluin1979finding,angluin1980inductive}, and its recent adaptation to the context of \emph{language generation} by \citet{kleinberg2024language}.
Specifically, we propose a novel theoretical abstraction to formally study the feasibility of reliably detecting hallucinations produced by language models. In our model, the hallucination detector is provided with a corpus of training data coming from some unknown target language $K$ and is allowed to interact with an LLM, whose outputs we denote by the set $G$. 
Conceptually, the language $K$ encodes all statements that are factually correct, while any output outside of $K$ is considered a \emph{hallucination}. We say that a hallucination detection algorithm is \emph{successful} if, after observing sufficiently many examples from $K$ and interacting extensively with the LLM, it eventually determines correctly whether or not the LLM produces hallucinations. Formally, this means that if $G \subseteq K$, the detector should eventually conclude that the LLM does not hallucinate, whereas if $G \not\subseteq K$, meaning the LLM generates outputs outside of $K$, the detector should correctly identify that the LLM hallucinates.

Our first main result formally establishes an equivalence between hallucination detection and the classical problem of language identification, which is known to be inherently challenging \citep{gold1967language,angluin1980inductive}. The practical implication is summarized concisely below:

\smallskip \begin{curvybox} \textbf{Informal Result I.}
Automated detection of hallucinations by a detector that is trained only on correct examples (positive examples) is inherently difficult and typically impossible without additional assumptions or signals. \end{curvybox} \smallskip

\noindent Thus, this result provides theoretical justification for the challenges encountered in practice when trying to automatically detect whether an LLM hallucinates.

Given this negative finding, we next examine a more optimistic scenario, inspired both by classical theory \citep{gold1967language} and modern empirical approaches \citep{kamoi2024evaluating}, in which the detector receives both correct statements (\emph{positive examples}) and explicitly labeled incorrect statements (\emph{negative examples}). Under these conditions, the outlook changes dramatically:

\smallskip \begin{curvybox} \textbf{Informal Result II.} Reliable automated hallucination detection is achievable when the detector is trained using both correct (positive) and explicitly labeled incorrect (negative) examples. \end{curvybox} \smallskip

This result has an interesting implication for practical attempts to create hallucination detectors: \emph{explicit expert feedback, particularly negative examples, is critical and fundamentally necessary for automated hallucination detection to succeed.}

These theoretical results align closely with both classical and recent theoretical findings, reinforcing the crucial importance of negative examples first noticed by \citet{gold1967language}, and complementing recent theoretical works on the benefits of negative examples for generating a broad set of responses without encountering hallucinations \citep{kalai2024calibrated,kalavasis2024characterizationslanguagegenerationbreadth,kalavasis2025limitslanguagegenerationtradeoffs}. They also resonate with current practical methodologies, such as reinforcement learning with human feedback (RLHF), that leverage explicit negative examples to reduce hallucinations and enhance model reliability in practice \citep{yang2024formal,deepmind2024alphaproof}.
In short, our theoretical findings provide a first theoretical understanding of the fundamental limitations—and necessary conditions—of automated hallucination detection in LLMs.

\subsection{Related Work}

\subsubsection{Theoretical frameworks for LLMs}
Our proposed framework builds on seminal works in learning theory, including the seminal Gold-Angluin framework \citep{gold1967language,angluin1979finding,angluin1980inductive} for language identification, and its recent adaptation to language generation by \citet{kleinberg2024language}. 

Following Kleinberg’s and Mullainathan's formulation, \citet{li2024generationlenslearningtheory} extended this perspective, using a learning-theoretic lens to characterize when ``uniform'' and ``non-uniform'' language generation are achievable. Further, recent works by \citet{kalavasis2025limitslanguagegenerationtradeoffs,kalavasis2024characterizationslanguagegenerationbreadth,charikar2024exploringfacetslanguagegeneration} explored notions of generation ``with breadth,'' demonstrating that this goal is inherently harder than standard language generation, and, in some cases, as challenging as language identification itself. 
In a similar spirit, \citet{peale2024} formalized
and studied a notion of ``representative generation,''
and showed that it is possible to achieve it in the limit 
for 
all countable collections of languages.
In a complementary direction, \citet{raman2025generation} studied language generation from \emph{noisy} examples, considering scenarios where training data includes instances outside the target language $K$.
In a concurrent and independent work, \citet{kleinberg2025densitymeasureslanguagegeneration} studied
a hallucination-breadth trade-off based on a notion of \emph{density} of languages.

Diverging from the Gold-Angluin framework, \citet{kalai2024calibrated} connected \emph{calibration} in generation to increased hallucination rates. Additionally, recent analyses by \citet{peng2024limitations,chen2024theoretical} identified fundamental limitations of transformer architectures. Using techniques from communication complexity, they proved transformers are incapable of composing functions when domains become sufficiently large, providing rigorous evidence for inherent hallucination tendencies in LLMs, given that function composition underlies reasoning \citep{guan2024mitigating}. Lastly, \citet{xu2024hallucination} leveraged complexity theory tools to demonstrate inevitable hallucinations in LLMs under certain assumptions. Earlier work by \citet{hanneke2018actively}
also illustrates the value of using external
feedback to mitigate hallucinations of generative models.
Our work contributes to this literature
which aims to give theoretical insights into the capabilities and limitations of LLMs.

\subsubsection{Empirical works on automated hallucination detection} 
Automated hallucination detection has recently gained significant attention, driven by the practical urgency to mitigate hallucinations. Several empirical approaches have emerged to tackle this challenge. For instance, \citet{manakul2023selfcheckgpt} introduce SelfCheckGPT, a black-box hallucination detection method that relies solely on stochastic sampling of model responses. The core intuition of their method is that factually accurate responses are typically consistent and frequent, whereas hallucinated outputs tend to vary and contradict each other.

In contrast to the black-box consistency-based method, \citet{azaria2023internal} propose leveraging the internal hidden states of the LLM to classify outputs as hallucinated or factual. Notably, their classifier is trained using an explicitly labeled dataset comprising sentences marked as either correct or incorrect, highlighting the benefit of explicitly supervised hallucination detection. Their results significantly outperform probability-distribution-based methods, illustrating the advantage of internal-state supervision and leveraging
annotated datasets to perform this task.

Building upon these empirical insights, \citet{kamoi2024evaluating} conduct a comprehensive evaluation demonstrating the limitations of current LLM-based hallucination detection approaches. In particular, they show that LLMs perform poorly as detectors when evaluating responses generated by other models, emphasizing the challenge in using LLMs for automated hallucination detection without robust external signals. Echoing this observation, \citet{tyen2023llms} further demonstrate that introducing even minimal human feedback greatly enhances the capability of LLMs to reliably detect hallucinations.
Similarly motivated, \citet{niu2023ragtruth} illustrate the benefits of fine-tuning LLMs using carefully curated, high-quality labeled datasets containing explicit annotations of hallucinations. This supervised fine-tuning improves hallucination detection performance and underscores the importance of explicitly labeled negative examples. 

Our theoretical findings provide formal validation for these empirical results, clearly highlighting the crucial role played by explicitly labeled negative examples in successful hallucination detection. 

For comprehensive surveys on the broad topic 
of hallucinations in LLMs, including various detection
methods discussed above, we refer the interested reader
to \citet{ji2023survey,zhang2023siren}.

\section{Model and Formal Results}

\subsection{Model}\label{sec:model}
In this section we define the formal model we consider in this work. 
We denote by $\cL = \inbrace{L_1, L_2, \ldots }$ a countable collection of candidate
languages, where each language $L_i$ is a subset of some countable domain $\cX.$
We assume that we have membership access to the collection $\cL$, meaning
that for any $i \in \N$ and $x \in \cX$ we can ask whether $x \in L_i.$ We allow
$\cL$ to contain multiple occurrences of the same language, \ie{} there might exist $i \neq j$ such that $L_i = L_j.$\footnote{This is because there might be different canonical representations of the same language.} Each language $L_i$ can have finite 
or infinite cardinality.\footnote{In the model of \citet{kleinberg2024language} the languages need to have infinite cardinality; this is not needed in our model.}
We define an enumeration of a language $L$ to be an infinite sequence $E = \inparen{w_1, w_2, w_3, \ldots}$ such that for all $i \in \N$ we have $w_i \in L$,
and for all $x \in L$ there is some $j \in \N$ such that $w_j = x.$ Notice that this
allows for repetitions of strings, but, crucially, for any given string $x \in L$
there is a finite index where this string appears. 

We define the hallucination detection game as the following interaction
between a learner and an adversary: the adversary picks a target language $K \in \cL$, an arbitrary enumeration $E = \inparen{w_1,w_2,\ldots}$ of $K$, and a target set $G \subseteq \cX.$ We say that $G$ \emph{hallucinates} with respect to $K$
if it contains elements outside of $K$, \ie, if $G \not\subseteq K$. 
We denote by $E_t = \inparen{w_1,\ldots,w_t}$ the prefix
of the first $t$ elements of $E.$
In every timestep $t = 1,2,\ldots,$ the learner observes $w_t$, asks
 \emph{finitely} many membership queries to $G,$ \ie{}, for finitely
many $x_1,\ldots,x_k \in \cX$ it can ask if $x_i \in G,$ and
get the correct response. Then, it has to output
its guess $g_t \in \inbrace{0,1}$ whether $G \subseteq K$; it outputs 0 if it believes $G$ hallucinates and 1 otherwise. We say that the learner detects hallucinations
in the limit if for every target language $K \in \cL,$ enumeration $E$ of $K$,
and candidate $G \subseteq \cX$ it holds that after sufficiently long but finite $t$ the guesses of the learner
become correct, \ie{}, there exists some $t_0 \in \N$ such that $g_t = \ind\inbrace{G \subseteq K}, \forall t \geq t_0.$ {The formal definition is provided below.}

     \begin{definition}
            [Hallucination Detection in the Limit]\label{def:hallucination-detection}
        Fix some $K$ from the language collection $\cL = \{L_1, L_2,\dots\}$ and some
            set $G \subseteq \cX$.
                The hallucination detection algorithm  $\algo{D} = (\algo{D}_t)$ detects hallucinations for $G$ in the limit if there is some $t^* \in \N$ such that for all steps $t > t^*$, the detector’s guess $d_t$ satisfies $d_t = \ind\inbrace{G \subseteq K}.$
                The language collection $\cL$ allows for hallucination detection in the limit if there is a hallucination detector that detects in the limit for any $K \in \cL,$ for any $G\subseteq\cX$, and for any enumeration $E$ of $K$.
        \end{definition}

To gain some intuition about this model, it is useful to consider a simple 
example.

\begin{example}\label{ex:multiples-of-naturals}
    Let $\cX = \N = \inbrace{1,2,3,\ldots}$ and $\cL = \inbrace{L_1, L_2, L_3,\ldots},$ where $L_i = \inbrace{i \cdot j, j\in \N},$ \ie{}, the 
    $i$-th language contains all the multiples of $i.$ Assume the adversary chooses $K = L_2,$ \ie{}, the language of all even numbers. Then, it has to present to the learner 
    all the even numbers, one at a time, potentially allowing for duplicates in the presentation. Crucially, for every even number $2\cdot j,$ there is some timestep $t_j \in \N$ such that $w_{t_j} = 2\cdot j.$
    Consider two potential choices of $G:$ the first choice
    is $G_1 = L_4$ and the second choice is $G_2 = L_3.$\footnote{We underline that we do not place the restriction $G \in \cL,$ this is only done to illustrate our example.} In the first
    case, a successful hallucination detection algorithm should claim, in the limit,
    that $G_1$ does not hallucinate with respect to $K$, whereas 
    in the second case it should claim that $G_2$ does hallucinate with
    respect to $K.$ 
    To give the reader a first glance of the difficulty of the hallucination
    detection task,
    while we have not stated our main result yet (\cref{thm:hallucination-detection-iff-identification}), it is worth pointing out 
    that this result, along with Angluin's characterization of language detection in the limit (\cref{thm:angluin-id-limit}), implies that no hallucination detection
    algorithm exists for this collection.
\end{example}

\paragraph{Identification and generation in the limit.} Our model 
is closely related to the Gold-Angluin language identification setting \citep{gold1967language,angluin1979finding,angluin1980inductive}, and the language
generation setting of \citet{kleinberg2024language}. In both of these models
there is a infinite game between a learner and an adversary: the adversary 
picks a target $K \in \cL$ and an enumeration of that target;  {{however, unlike these models, the adversary does not pick
a target set} $G \subseteq \cX$} as happens in our setting. {{Similar to these models}}, in every $t \in \N$
the learner observes a new element from the enumeration. In the identification setting,
the goal of the learner is to find an index of the target language $K$ for 
all but finitely many steps, and in the generation setting the goal is to output
\emph{unseen} elements of $K$ for all but finitely many steps.
We present a more
formal treatment of these settings in \cref{app:preliminaries}. \citet{angluin1980inductive} exactly characterized when identification in this setting
is achievable. Her result is largely viewed as an impossibility result, since a very limited number of collections satisfy it. On the other hand,
\citet{kleinberg2024language} showed that the landscape of generation is vastly different: 
it is achievable for all such collections. 

\paragraph{Connections of theoretical model to practical LLM training.} At this point, it is instructive to pause and consider some common features
of these three models; we believe that while they are mathematical
abstractions of the practical LLM training process, they capture 
a lot of important aspects of this process. 
The way to interpret the different languages of the collection
$\cL$ is that they capture different ``worlds'' and the different
elements of $\cX$ are different ``statements.'' 
Therefore, each ``world'' defines
precisely which ``statements'' are accurate and which ones are inaccurate. Admittedly, real-world applications might be more nuanced than that
and there could be statements that cannot be easily categorized
into accurate or inaccurate ones. Since our model gives a clear
taxonomy, it follows that a negative result here can be viewed
as a strong indication that in real-world applications hallucination
detection is even more challenging. 

In our model, we consider an \emph{adversarial enumeration} instead
of placing distributional assumptions on the language generation process
and the way the outputs of the LLM are generated.
While this might look like a restriction of our model at first glance, 
it turns out that our results carry over to a setting where the data are
generated probabilistically; this follows from techniques 
similar to \citet{kalavasis2025limitslanguagegenerationtradeoffs}. 
We choose to focus on the adversarial setting, following the work of 
\citet{kleinberg2024language}, to make our exposition easier to follow. 

Moreover, in all these models the ``learner'' is given access only
to \emph{positive examples} in the form of elements that belong to the
target language. This assumption is capturing the pre-training
process of modern machine learning architectures that are trained on a large corpus of datapoints
that are elements of the target language {and are deployed to act at automatic language identifiers, generators or detectors}. We also ignore the fact that the training dataset might be corrupted. Again, this simplification 
is made to ensure that our negative result reflects an innate difficulty of the hallucination detection task and is not an artifact of inaccuracies
contained in the training data.

Next, notice that in all three settings we have discussed so far -- language identification, language generation, and hallucination detection -- the algorithm never receives feedback about its guesses. This is also
largely consistent with the pre-training phase of the LLM training pipeline.

Furthermore, we do not place any computational restrictions on the learning
algorithm or the architecture that it relies upon. In fact, we only
wish for the detection property to hold ``in the limit.'' This simplification
is again made to ensure that any negative results in the setting
reveal inherent difficulties of the underlying task and are not 
mere limitations of the current technologies or computational resources
that might be rectified in the future.

Lastly, we underline that throughout our work we consider a 
\emph{promptless} generation setting. Intuitively, this is also 
a simplification of the behavior of real-world LLMs, thus negative results
in our model should also carry over to applied settings. We emphasize
that all these assumptions are largely consistent with prior
work on theoretical capabilities and limitations of LLMs  \citep{kalai2024calibrated, xu2024hallucination, kleinberg2024language, kalavasis2024characterizationslanguagegenerationbreadth, charikar2024exploringfacetslanguagegeneration, kalavasis2025limitslanguagegenerationtradeoffs}. We believe that our
results can be extended to the prompted setting of \citet{kleinberg2024language}, and we leave this as an interesting future
direction.

\subsection{Formal Results}\label{sec:results}
Given the similarities of the different tasks we have described so far---language identification, language generation, and hallucination detection---it is natural to ask: is hallucination detection
as easy as generation, as hard as identification, or does it lie somewhere in between? 
Our first main result gives a precise answer to this question; we show that hallucination
detection is as hard as identification. 

\begin{theorem}\label{thm:hallucination-detection-iff-identification}
    A countable collection of languages $\cL = \inbrace{L_1,L_2,\ldots}$ over some countable domain
    $\cX$ admits an algorithm that detects hallucinations in the limit if and only if $\cL$ is 
    identifiable in the limit.
\end{theorem}

Given our result and Angluin's characterization \citep{angluin1980inductive} which
we state in \cref{thm:angluin-id-limit}, we get the following immediate corollary.

\begin{corollary}\label{cor:}
     A countable collection of languages $\cL = \inbrace{L_1,L_2,\ldots}$ over some countable domain
    $\cX$ admits an algorithm that detects hallucinations in the limit if and only if $\cL$
    satisfies Angluin's condition (\cref{def:angliun-criterion}).
\end{corollary}

Given this largely negative result about the ability to perform automated hallucination detection
of LLMs, we next ask what more information is needed by the learner to perform this task. Inspired
by Gold's work \citep{gold1967language}, we consider a modified game termed hallucination detection
with negative examples. The main difference is that instead of presenting an enumeration of the
target language $K$, the adversary presents an enumeration of the whole domain $\cX$ along with a label in $\inbrace{0,1}$
indicating whether the enumerated element is in the target language or not. {We call this type of enumeration a \emph{labeled} enumeration.} In stark
contrast to our previous result, we show that hallucination detection with negative examples
is always possible. The formal description of this game i{given below}.


   {     \begin{definition}
            [Hallucination Detection with Negative Examples in the Limit]\label{def:hallucination-detection-negative-examles}
                  Fix some $K$ from the language collection $\cL = \{L_1, L_2,\dots\}$ and some
            set $G \subseteq \cX$.
                The hallucination detection algorithm  $\algo{D} = (\algo{D}_t)$ detects hallucinations for $G$ given negative examples in the limit if there is some $t^* \in \N$ such that for all steps $t > t^*$, the detector’s guess $d_t$ satisfies $d_t = \ind\inbrace{G \subseteq K}.$
                The language collection $\cL$ allows for hallucination detection with negative examples in the limit if there is a hallucination detector that detects in the limit for any $K \in \cL,$ for any $G\subseteq\cX$, and for any labeled enumeration $E$ of $\cX$ with respect to the target language $K$.
        \end{definition}

        Having explained the mathematical setting, we are now ready to state our formal result.
}

\begin{theorem}\label{thm:detection-negative-examples}
    Every countable collection of languages $\cL = \inbrace{L_1,L_2,\ldots}$ over some countable domain
    $\cX$ admits an algorithm that, given negative examples, detects hallucinations in the limit.
\end{theorem}

\section{Overview of the Approach}\label{sec:overview-of-approach}
Having discussed our formal setting and results, we now describe
the main steps of our technical approach. We start with \cref{thm:hallucination-detection-iff-identification}.

\subsection{Proof of \cref{thm:hallucination-detection-iff-identification}}
Our approach here is divided into two main steps. First, we show
that we can transform any algorithm that achieves identification
in the limit in this setting to an algorithm that detects
hallucinations in the limit. 

\paragraph{Language identification $\implies$ hallucination detection.}
The formal statement of this result is given below. 

\begin{lemma}\label{lem:identification-implies-hallucination-detection}
    Let $\cL$ be a countable collection of languages over a domain $\cX$
    that is identifiable in the limit. Then, $\cL$ admits an algorithm
    that achieves hallucination detection in the limit.
\end{lemma}

Let us now explain the idea of our approach, which utilizes the identification algorithm in a black-box way. In every timestep $t$, the learner feeds the element $w_t$ the adversary enumerates to the identification algorithm.
The identification property (\cref{def:Identification}) immediately
shows that there exists some $t^* \in \N$ (which depends on the choice of the target language $K$ and the enumeration $E$) 
such that for all $t > t^*$
the identifier's guess $i_t$ satisfies $i_t = i_{t-1}$ and $L_{i_t} = K.$
The learner next considers an enumeration of the domain $\cX = \inbrace{x_1, x_2,\ldots}$. 
In the $t$-th step of the execution, the learner uses the membership
oracle to check which of the elements $x_1,\ldots,x_t$ belong
to the language $L_{i_{t}}.$ Subsequently, the learner also queries the target LLM, modeled as the set $G$, to see which of these elements belong to it.
If for all $x_i \in G$ it holds that $x_i \in L_{i_t},$ then
the guess of the hallucination detection algorithm for this step
will be that the LLM does not hallucinate. We present a formal overview
of our hallucination detection strategy in \cref{alg:ident2halluc}. We now 
give the formal proof of our result.

{
 \begin{algorithm}[H]
\caption{Hallucination Detection from Language Identification}
\label{alg:ident2halluc}
\begin{algorithmic}[1]

\algrenewcommand\algorithmicrequire{\textbf{Input:}}
\Require
\Statex \quad $\bullet$ Identification algorithm $\algo{I}$
\Statex \quad $\bullet$ Enumeration $E=(w_1, w_2,\ldots)$ of $K$
\Statex \quad $\bullet$ Language collection $\cL$
\Statex \quad $\bullet$ Domain $\cX$
\Statex \quad $\bullet$ Membership oracle for $\cL$
\Statex \quad $\bullet$ LLM output set $G$

\For{$t = 1, 2, \dots$}
    \State Feed $E_t = \bigl(w_1,\ldots,w_t\bigr)$ to $\algo{I}$ to obtain guess $i_t$.
    \State Let $\hat K \leftarrow L_{i_t}$. 
    \State Enumerate domain prefix $\cX_t = \{x_1,\dots,x_t\}$.
    \For{each $x \in \cX_t$}
        \If{$x \in G$ and $x \notin \hat K$}
            \State \Return $G$ hallucinates
        \EndIf
    \EndFor
    \State \Return $G$ does not hallucinate
\EndFor
\end{algorithmic}
\end{algorithm}
}

\begin{proof}[Proof of \cref{lem:identification-implies-hallucination-detection}]
First, notice
that by definition of the identification property, it holds
that there exists some $t^* \in \N$ (that depends both on the target
language and the enumeration) such that for all $t \geq t^*$
the output of the identifier satisfies $L_{i_t} = K.$ Let us 
now consider any $t > t^*.$ We divide our analysis into two disjoint
cases, which jointly cover all possible outcomes.
\begin{itemize}
    \item First, let us consider the case $G \subseteq K.$ Then,
    for all $t \geq t^*$ we have that if $x_i \in G$ then $x_i \in K,$
    for all $1 \leq i \leq t.$ Thus, our algorithm will correctly
    claim that the LLM does not hallucinate for all $t \geq t^*.$

    \item Next, we consider the slightly more challenging case
     $G \not\subseteq K.$ By definition, there exists some 
    $x \in \cX$ such that $x \in G$ and $x \not\in K.$ Let 
    $i^* \in \N$ be the smallest index in the enumeration of $\cX$
    for which this holds, \ie{}, $x_{i^*} \in G, x_{i^*} \not\in K.$ Then, for any $t \geq \max \inbrace{t^*, i^*}$
    when we consider the prefix of the enumeration $x_1,\ldots,x_t$
    the element $x_{i^*}$ will be included in this enumeration. 
    Moreover, it holds that $L_{i_t} = K.$ Thus, when the hallucination
    detector tests the element $x_{i^*}$, it will see that 
    $x_{i^*} \in G$ and $x_{i^*} \not\in K$ and it will correctly
    deduce that the LLM $G$ hallucinates.    
\end{itemize}

The previous two arguments conclude the proof.
\end{proof}

\paragraph{Hallucination detection $\implies$ language identification.}
We now shift our attention
to the more technically intricate result which shows that language
identification is not harder than hallucination detection. This is 
also a black-box transformation; it takes as input any hallucination
detection algorithm and it constructs an identification algorithm.

\begin{lemma}\label{lem:hallucination-detection-implies-identification}
    Let $\cL$ be a countable collection of languages over a domain $\cX$
    that admits an algorithm that achieves hallucination detection in the limit.
    Then, $\cL$ is identifiable in the limit.
\end{lemma}

Before explaining our construction, it is instructive to build 
some intuition about the difficulty of the language identification task.
 A natural attempt
to achieve language identification is to keep track of all the
language that are ``consistent'' with the current set of 
examples $E_t$ the adversary has enumerated,
that is the set $\cC_t = \inbrace{L \in \cL: E_t \subseteq L}.$
It is not very hard to see that for any language $L_i$ that is not a (strict)
superset of the target language $K$, there is some timestep $t^*_i$ such that
$L_i \notin \cC_t.$ Indeed, since $L_i \not\supset K,$ there exists
some $x_i$ which satisfies $x_i \in K, x_i \notin L_i.$
Thus, when the adversary enumerates $x_i$ the algorithm will deduce
that $L_i \not\in \cC_t.$ What happens if $L_i$ is a (strict) superset of
$K$? Unfortunately, in this case the language $L_i$ will always remain
consistent with the sample $E_t.$ Thus, the strategy of 
keeping track of the consistent languages is not sufficient to guarantee
identification in the limit. Indeed, consider \cref{ex:multiples-of-naturals}: no matter what the choice the target language 
$K$ and the enumeration $E$ of the adversary is, the
language $L_1$ will always be consistent with the sample $E_t.$
Thus, the consistency-based approach is not sufficient to distinguish
between $L_j, j\neq 1,$ and $L_1.$ Is there a more sophisticated approach
that can overcome this obstacle? The seminal result of
\citet{angluin1980inductive} shows that, unless $\cL$ satisfies
some very strong structural conditions (\cref{def:angliun-criterion}),
the answer is largely negative.

The previous discussion highlights that in order to achieve identification
in the limit we need to leverage the hallucination detection algorithm
to distinguish between languages $L_i$ with $L_i \supsetneq K$ and the target language $K.$ Our main insight is that the ``consistency-based''
approach and the hallucination detection algorithm work in a complementary way: the former allows us to discard languages that are not (strict) supersets of $K$, while the latter helps us rule out languages that are (strict) supersets of $K.$\footnote{In fact, the hallucination detection algorithm allows us to discard languages that are \emph{not} subsets of $K.$} Neither of these approaches is sufficient to be used for
language identification on its own, but it turns out that a carefully
crafted approach that coordinates their behavior gives the desired result.

We now explain our algorithm in more detail; its formal description is given
in \cref{alg:halluc2ident}. 
In each step $t$ we create the set $\cC'_t = \inbrace{L_i \in \cL: E_t \subseteq L_i, 1 \leq i \leq t},$ \ie{}, the set of the languages whose
index is at most $t$ and are consistent with the elements $E_t$
that have been enumerated so far. Notice that this can be achieved
with finitely many queries to the membership oracle for $\cL.$\footnote{In fact, we only need $2t-1$ fresh queries in the $t$-th round.} 
Let $L_{i_1},\ldots,L_{i_k}$ be the languages of $\cC'_t.$
Next, we 
run $k$ copies of the hallucination detection algorithm: the $i$-th
copy is given as input the collection $\cL,$ the currently enumerated
set $E_t$, and the target set $L_i$ as the LLM that needs to be tested for hallucinations. Our guess for the target language
is the \emph{smallest} element $z'$ for which \textbf{i)} $L_{z'} \in \cC'_t$, and \textbf{ii)} the output of the $z'$-th copy of the hallucination detection algorithm guesses that $L_{z'}$ does not
hallucinate. We now give the formal proof.

{
\begin{algorithm}[H]
\caption{Identification via Hallucination Detection}
\label{alg:halluc2ident}
\begin{algorithmic}[1]

\Require
\Statex \quad $\bullet$ Hallucination detection algorithm $\algo{D}$
\Statex \quad $\bullet$ Enumeration $E=(w_1, w_2,\dots)$ of the target language $K$
\Statex \quad $\bullet$ Language collection $\cL$
\Statex \quad $\bullet$ Domain $\cX$
\Statex \quad $\bullet$ Membership oracle for $\cL$

\For{$t = 1, 2, \dots$}
    \State Let $E_t = (w_1, w_2, \dots, w_t)$.
    \State Compute the \emph{consistent set}
    \[
      \cC'_t = \{L_i \in \cL : E_t \subseteq L_i \text{ and } i \le t\}.
    \]
    \For{$i=1,\ldots,t$}
        \State Run a copy of $\algo{D}$ with inputs $E_t$ and target set $L_i$ for $t$ steps, and obtain output $d_t^i$, where:
        \[
        d^t_i= 
          \begin{cases}
          1 & \text{if no hallucinations are detected,}\\[1mm]
          0 & \text{if hallucinations are detected.}
          \end{cases}
        \] 
    \EndFor
    \State Let 
    \[
      \cN = \{ i \le t : L_i \in \cC'_t \text{ and } d^i_t=1 \}.
    \]
    \If{$\cN \neq \emptyset$}
        \State Let $z' = \min\inbrace{i \in \cN}$.
        \State \Return ${z'}$ \Comment{Output the index of the identified language.}
    \Else
        \State \Return 1. \Comment{We return an arbitrary index and proceed.}
    \EndIf
\EndFor

\end{algorithmic}
\end{algorithm}
}


\begin{proof}[Proof of \cref{lem:hallucination-detection-implies-identification}]
We let $z \in \N$ be the smallest number such that $L_z = K.$\footnote{Recall that a language is allowed to appear multiple times in the collection $\cL.$} Our algorithm outputs the language with the \emph{smallest} index that satisfies these two conditions we described above. Thus, 
to get the desired result we need to show that \textbf{i)} all the languages
in $\cL_{z-1} = \inbrace{L_1, L_2, \ldots, L_{z-1}}$
that precede $L_z$ do \emph{not} satisfy
these conditions (for all sufficiently large $t$), while the target 
language $L_z$ \emph{does} satisfy these conditions (again, for all sufficiently large $t$). To that end, we divide $\cL_{z}$ into two disjoint subsets:
$\cL^{\supset}_{z-1} = \inbrace{L \in \cL_{z-1}: L \supsetneq L_z}$ and
$\cL^{\not\supset}_{z-1} = \inbrace{L \in \cL_{z-1}: L \not\supsetneq L_z}$.
In words, $\cL^{\supset}_{z-1}$ is the set of all languages that
precede $L_z$ and are strict supersets of it, and $\cL^{\not\supset}_{z-1}$
is the set of all languages that
precede $L_z$ and are \emph{not} strict supersets of it.
Notice that, since $L_z \not\in \cL_{z-1}$, we have $\cL^{\supset}_{z-1} \cup \cL^{\not\supset}_{z-1} = \cL_{z-1}.$ We now handle these two sets separately.

We first consider the set $\cL^{\not\supset}_{z-1} = \inbrace{L \in \cL_{z-1}: L \not\supsetneq L_z}$. We denote by $L_{i_1},\ldots, L_{i_k}$ the languages of this collection, where $0 \leq k \leq z-1.$ By definition, for every such $L_{i_j}$
there exists some element $x_{i_j} \in L_z, x_{i_j} \not\in L_{i_j}.$ Moreover, 
since the adversary presents a complete enumeration of $L_z$
there exists some timestep $t_{\ell_j}$ such that $w_{t_{\ell_j}} = x_{i_j}$ (recall that $w_{t_{\ell_j}}$ is the element enumerated by the adversary at timestep $t_{\ell_j}$.)
We define $t^*_1 = \max_{ j \leq k} t_{\ell_j}.$ Using the definition
of the consistent set $\cC'_{t}$ we see that for all $t\geq t^*_1$
these languages are not consistent with $E_t$, \ie, $L_{i_1}, \ldots, L_{i_k} \not\in \cC'_{t}.$

We now focus on the set $\cL^{\supset}_{z-1}.$ Let $L_{j_1},\ldots,L_{j_m},$
be the languages of this collection, where $0 \leq m \leq z-1.$ For any $i\leq m$ consider
the execution of the hallucination detection algorithm with input the collection $\cL,$ the enumeration $E,$ and the target set $L_{j_i}$.
Since $E$ is a valid enumeration of $L_z$ and $L_{j_i} \supsetneq L_z,$
by definition of the hallucination detection property, there exists
some $t'_{\ell_i}$ such that for all $t \geq t'_{\ell_i}$ the hallucination
detection algorithm declares that $L_{j_i}$ hallucinates. To see that, notice that since $L_{j_i} \supsetneq L_z$ and
the hallucination detection algorithm observes a sequence that enumerates all of $L_z,$ it must eventually conclude that 
$L_{j_i}$ hallucinates.
We define $t^*_2 = \max_{i\leq m} t'_{\ell_i}.$ It follows that
for all $t \geq t^*_2$ the hallucination detection algorithm
declares that each $L_{j_1},\ldots,L_{j_m}$ hallucinates.

Lastly, let us consider the language $L_z.$ First, notice that for all
$t \geq z$ we have $L_z \in \cC'_t.$ Moreover, using the exact similar
reasoning as in the above paragraph, there exists some timestep
$t'$ such that for all $t \geq t'$ the hallucination detection algorithm
declares that $L_z$ does not hallucinate. Let $t^*_3 = \max\inbrace{z, t'}.$

We now have all the ingredients we need to prove our result. 
We let $t^* = \max\inbrace{t^*_1,t^*_2,t^*_3}.$ Consider any $t \geq t^*.$
By definition of $t^*$, 
for any language $L_i, i <z,$ either $L_i \not\in \cC'_t$
or the hallucination detection algorithm with input set $L_i$ declares
that $L_i$ hallucinates, so the two conditions are not simultaneously satisfied for this language. However, both conditions are simultaneously satisfied for $L_z$ since $L_z \in \cC'_t$ and the hallucination detection algorithm declares that $L_z$ does not hallucinate.
Hence, the smallest indexed language that satisfies both of our conditions
is indeed $L_z.$ Consequently, our algorithm achieves identification in the limit.
\end{proof}

We now note that the proof of \cref{thm:hallucination-detection-iff-identification}
follows as an immediate corollary of \cref{lem:identification-implies-hallucination-detection,lem:hallucination-detection-implies-identification}.

\subsection{Proof of \Cref{thm:detection-negative-examples}}
Unlike \cref{thm:hallucination-detection-iff-identification}, the
technical details of \cref{thm:detection-negative-examples} 
are not as challenging. The full proof of this result is given below.

\begin{proof}[Proof of \cref{thm:detection-negative-examples}]
    We first describe the strategy we use to achieve hallucination
    detection. Recall that $G$ denotes the set we test, $K$ denotes
    the target language, and $E = \inbrace{\inparen{w_1,y_1}, \inparen{w_2,y_2}, \ldots},$ where $y_i = \ind\inbrace{w_i \in K}, \forall i\in \N,$ in a labeled enumeration of $\cX$, \ie, every element
    $x \in \cX$ appears at some finite position in the enumeration, and
    its label indicates whether it is part of the target language $K.$
    In every step $t=1,2,\ldots,$ we do the following: 
    \begin{itemize}
        \item For every element in the input stream that appears with a 0 label, \ie, $(w,0) \in E_t \coloneqq \inbrace{\inparen{w_1,y_1},\ldots,\inparen{w_t,y_t}}$ 
        we check if $\ind\inbrace{w \in G} = 1.$ If this holds
        for some $(w,0) \in E_t$ we declare that $G$  hallucinates.

        \item Otherwise, we declare that $G$ does not hallucinate.
    \end{itemize}
    We now prove the correctness of this strategy. Similarly as before,
    we divide the analysis into two cases.
    \begin{itemize}
        \item If $G \subseteq K,$ then the above algorithm 
        correctly declares that $G$ does not hallucinate in every step
        $t \in \N.$ This is because $w \not\in K \implies w \not\in G.$
        \item If $G \not\subseteq K,$ we consider an enumeration of the domain $\cX = \inbrace{x_1,x_2,\ldots}.$ Let $i^* \in \N$ be the smallest
        number such that $x_{i^*} \in G$ and $x_{i^*} \not\in K.$ Notice
        that such an $i^*$ does exist. Moreover, there exists 
        some $t^*$ such that $\inparen{w_{t^*}, y_{t^*}} = \inparen{x_{i^*},0}.$ Thus, for this tuple we
        get $y_{t^*} = 0,$ and $\ind\inbrace{w_{t^*} \in G} = 1.$ Hence, for any $t \geq t^*$
         our algorithm will correctly declare that $G$ hallucinates.
    \end{itemize}
    This concludes the proof.
\end{proof}

\section{Conclusion}
In this work, we initiated the formal study of automated hallucination detection by introducing a mathematical framework to explore the possibilities and inherent limitations of this task. Our results provide theoretical justification for several phenomena observed experimentally. Specifically, we showed that hallucination detection is typically unattainable if detectors are trained solely on \emph{positive} examples from the target language (\ie, factually correct statements). In stark contrast, when detectors have access to explicitly labeled \emph{negative} examples—factually incorrect statements—hallucination detection becomes tractable for all countable collections. These findings underscore the critical role of human feedback in practical LLM training.
Several compelling directions for future work remain open. It would be valuable to quantify precisely the amount of negative examples needed for reliable hallucination detection, and formally explore the computational complexity of the detection problem within our proposed framework. Additionally, investigating whether hallucination detection remains tractable under noisy negative examples, as well as exploring alternative forms of feedback beyond explicit labeling, are promising avenues that warrant further exploration.
Finally, inspired by the definition of \citet{kleinberg2025densitymeasureslanguagegeneration} it would be interesting to explore whether we can achieve a more relaxed notion of hallucination detection, where we only wish
to detect whether the density of hallucinations is greater than some target threshold $c > 0.$

\section*{Acknowledgments}
We are grateful to Evgenios M. Kornaropoulos for pointing out and helping us correct an error in the proof of \Cref{thm:detection-negative-examples}.

\newpage

 \printbibliography
\newpage
\appendix
\section{Preliminaries}\label{app:preliminaries}
Building on the foundational work in learning theory by \citet{gold1967language} and \citet{angluin1988identifying}, \citet{kleinberg2024language} introduced a rigorous framework for language generation. In this model, the domain $\cX$ is a countable set, and the target language $K$ is an unknown subset of $\cX$.

    \subsection{Language Identification in the Limit}
    The problem of language identification in the limit from positive examples was introduced by \citet{gold1967language} and further studied by \citet{angluin1979finding,angluin1980inductive}.
    For a fixed collection $\cL$, an adversary and an identifier play the following game: 
    The adversary chooses a language $K$ from $\cL$ without revealing it to the identifier, and it begins \emph{enumerating} the strings of $K$ (potentially with repetitions) $w_1,w_2,\dots$ over a sequence of time steps $t = 1,2,3,\dots$. 
    The adversary can repeat strings in its enumeration,
    but the crucial point is that for every string $x \in K$, there must be at least one time step $t$ at which
    it appears. At each time $t$, the identification algorithm $\algo{I}$, given the previous examples $w_1,w_2,\dots,w_t$, outputs an index $i_t$ that corresponds to its guess for the index of the true language $K$.
        Language identification in the limit is then defined as follows.
        \begin{definition}
            [Language Identification in the Limit \citep{gold1967language}]\label{def:Identification}
            Fix some $K$ from the language collection $\cL = \{L_1, L_2,\dots\}$.
                The identification algorithm  $\algo{I} = (\algo{I}_t)$ identifies $K$ in the limit if there is some $t^* \in \N$ such that for all steps $t > t^*$, the identifier’s guess $i_t$ satisfies $i_{t} = i_{t-1}$ and $L_{i_t} = K.$
                The language collection $\cL$ is identifiable in the limit if there is an identifier that identifies in the limit any $K \in \cL,$ for any enumeration of $K$.
                {In this case, we say that the identifier identifies the collection $\cL$ in the limit.}
        \end{definition}
        Angluin's seminal result \citep{angluin1980inductive} proposed a condition that precisely characterizes which collections are identifiable in the limit.
        \begin{definition}
        [Angluin's Condition \citep{angluin1980inductive}]
        \label{def:angliun-criterion}
            Fix a language collection $\cL = \{L_1, L_2, \dots\}$.
            The collection $\cL$ is said to satisfy Angluin's condition if for any index $i$, there is a tell-tale, \ie{}, a finite set of strings $T_i$ such that $T_i$ is a subset of $L_i$, \ie{}, $T_i\subseteq L_i$, and the following holds:
            \begin{center}
                \centering 
                For all $j\geq 1$, if $L_j\supseteq T_i$, then $L_j$ is not a proper subset of $L_i$.
            \end{center}
            Further, the tell-tale oracle is a primitive that, given an index $i,$ outputs an enumeration of the set $T_i$.
        \end{definition}
       Formally, \citet{angluin1980inductive} showed the following result.
        \begin{theorem}
        [Characterization of Identification in the Limit \citep{angluin1980inductive}]
        \label{thm:angluin-id-limit}
            A language collection $\cL$ is identifiable in the limit if and only if it satisfies Angluin's condition. %
        \end{theorem}
        Perhaps surprisingly, this result shows that language identification is impossible even for simple collections.

    \subsection{Language Generation in the Limit}
       Using the same game-theoretic setting as \citet{gold1967language}, \citet{kleinberg2024language} proposed
       a modification of this game where the objective of the learner is to \emph{generate} unseen
       elements of $K$ instead of guessing its index.
        \begin{definition}
                [Language Generation in the Limit \citep{kleinberg2024language}]\label{def:consistentGeneration}
                Fix some $K$ from the language collection $\cL = \{L_1, L_2,\dots\}$ and a generating algorithm $\generator ~{ =\inparen{\generator_t}}.$
                    At each step $t$, let $E_t \subseteq K$ be the set of all strings that the algorithm $\generator$ has seen so far. 
                    $\generator$ must output a string $w_t \notin E_t$ (its guess for an unseen string in $K$). 
                    The algorithm  $\generator$ {is said to generate} from $K$ in the limit if, for all enumerations of $K$, there is some $t^* \in \N$ such that for all steps $t \geq t^*$, the algorithm’s guess {$w_t$} belongs to $K \setminus E_t$ {(or $K \setminus E_t$ is empty)}. The collection $\cL$
                    allows for generation in the limit if there is an algorithm  $\generator$ that, for any target $K \in \cL,$ 
                     generates from $K$ in the limit.
                \end{definition}
        Note that for the problem of language generation to be interesting, the languages of the collection $\cL$ must be of infinite cardinality.
        The main result of \citet{kleinberg2024language} is that language generation in the limit is possible for all countable collections of languages.
         \begin{theorem}
    [Theorem 1 in \citet{kleinberg2024language}]
    There is a generating algorithm with the property that for any countable collection of languages $\cL =
    \{L_1, L_2,\dots\}$, any target language $K \in \cL,$ and any enumeration of $K$, the algorithm generates from $K$ in the limit.
     \end{theorem}

\end{document}